\documentclass[letterpaper]{article} 
\usepackage{aaai25}  
\usepackage{times}  
\usepackage{helvet}  
\usepackage{courier}  
\usepackage[hyphens]{url}  
\usepackage{graphicx} 
\usepackage{natbib}  
\usepackage{caption} 
\usepackage{algorithm}
\usepackage{algorithmic}

\usepackage{cuted}
\usepackage{amsfonts}       
\usepackage{xcolor}         
\usepackage{nicefrac}       
\usepackage{amsmath}
\usepackage{colortbl} 
\usepackage{pifont}
\usepackage{lipsum}
\usepackage{url}            
\usepackage{booktabs}       
\newcommand{\xmark}{\ding{55}}
\usepackage{newfloat}
\usepackage{listings}
\DeclareCaptionStyle{ruled}{labelfont=normalfont,labelsep=colon,strut=off} 
\floatstyle{ruled}
\newfloat{listing}{tb}{lst}{}
\floatname{listing}{Listing}

\definecolor{Gray}{gray}{0.85}

\newcolumntype{a}{>{\columncolor{Gray}}c}

\title{
   \begin{minipage}{0.125\textwidth}
       \includegraphics[width=0.85\textwidth]{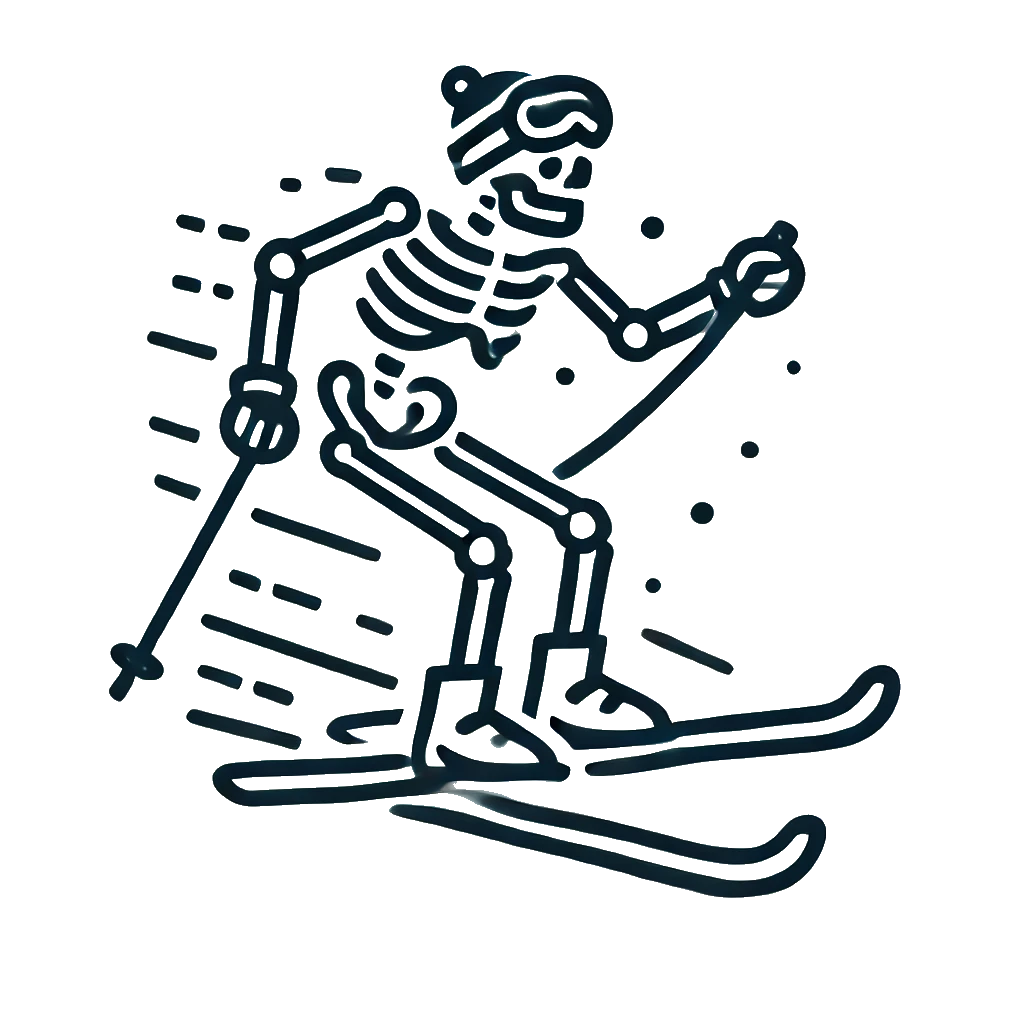}
   \end{minipage}
   \hspace{-0.04\textwidth}
   \begin{minipage}{0.85\textwidth}
       \centering
\textcolor{red}{SKI} Models: \textcolor{red}{SK}eleton \textcolor{red}{I}nduced Vision-Language Embeddings for Understanding Activities of Daily Living
   \end{minipage}
}

\author {
    Arkaprava Sinha\textsuperscript{\rm 1},
    Dominick Reilly\textsuperscript{\rm 1},
    Francois Bremond\textsuperscript{\rm 2, 3},
    Pu Wang\textsuperscript{\rm 1},
    Srijan Das\textsuperscript{\rm 1}
}
\affiliations {
    \textsuperscript{\rm 1}University of North Carolina at Charlotte \\
    \textsuperscript{\rm 2}Université Côte d'Azur \\
    \textsuperscript{\rm 3}Inria \\
    \{asinha13, sdas24\}@charlotte.edu
}

\begin{document}


\twocolumn[{
   \renewcommand\twocolumn[1][]{#1}%
   \maketitle
   \vspace{-5pt}
   \centering
   \scalebox{0.85}{
   \captionsetup{type=figure}
   \scalebox{0.9}{
   \includegraphics[width=1.25\textwidth, height=6.8cm]{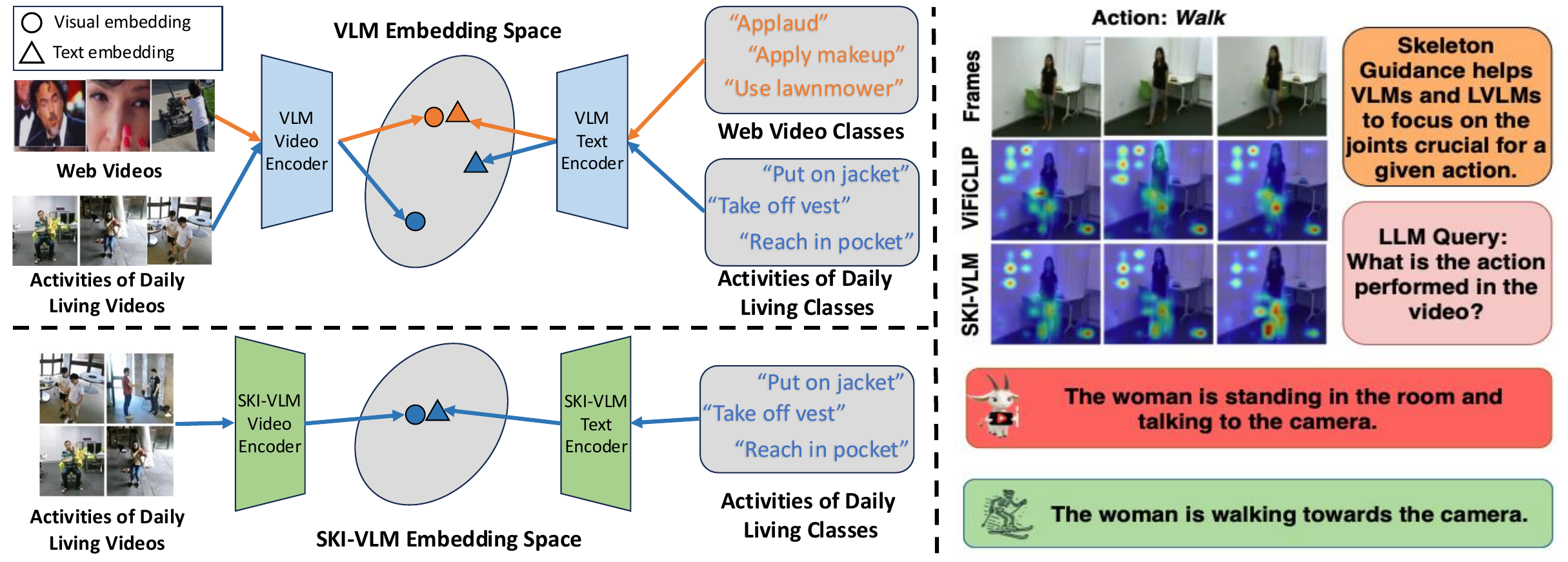}}
}

\setcounter{figure}{0} \vspace{-0.1in}
   \captionof{figure}{\textbf{Left}: The illustration depicts an embedding space of a Vision-Language Model (VLM) where representations of web-based videos align closely with their corresponding class label text features, while those of Activities of Daily Living (ADL) videos remain distant. Our study reveals that integrating skeleton guidance bridges this gap, aligning ADL video representations with their respective class labels. \textbf{Right}: Activation maps demonstrate how skeleton guidance sharpens the model’s focus on the critical body parts (such as legs) for specific actions, like \textit{Walk}. This enhancement is evident in the improved text descriptions generated by Large-Vision-Language Models (LVLMs) when queried about actions depicted in the videos.
   }\vspace{1em}
   \label{fig:teaser}
    \vspace{-0.035in}
}]

\begin{abstract}
The introduction of vision-language models like CLIP has enabled the development of foundational video models capable of generalizing to unseen videos and human actions. However, these models are typically trained on web videos, which often fail to capture the challenges present in Activities of Daily Living (ADL) videos. Existing works address ADL-specific challenges, such as \textit{similar appearances}, \textit{subtle motion patterns}, and \textit{multiple viewpoints}, by combining 3D skeletons and RGB videos. However, these approaches are not integrated with language, limiting their ability to generalize to unseen action classes.
In this paper, we introduce \textbf{SKI models}, which integrate 3D skeletons into the vision-language embedding space. SKI models leverage a skeleton-language model, \textbf{SkeletonCLIP}, to infuse skeleton information into Vision Language Models (VLMs) and Large Vision Language Models (LVLMs) through collaborative training. Notably, SKI models do not require skeleton data during inference, enhancing their robustness for real-world applications. The effectiveness of SKI models is validated on three popular ADL datasets for zero-shot action recognition and video caption generation tasks. Our code is available at \url{https://github.com/thearkaprava/SKI-Models}
\end{abstract}

\section{Introduction}
\label{sec:intro}
In recent years, the introduction of CLIP~\cite{CLIP}, has established the beneficial impact of language supervision in the training of vision-based discriminative models. In the video domain many works~\cite{vificlip, FROSTER, XCLIP} have extended the effectiveness of CLIP to video representation learning. 
These approaches have made promising strides especially in zero shot video action recognition where the target action labels are unavailable during training. A majority of these models have been 
trained on web-based videos~\cite{kinetics, ucf, kuehne2011hmdb} 
consisting of sports videos, movie clips, etc. These videos generally contain actions with prominent motion and are typically appearance based, aligning actions closely to their scenes; for instance, \textit{playing \textbf{soccer} on \textcolor{green}{green} grass} or \textit{\textbf{swimming} in \textcolor{blue}{blue} water}.
In contrast, ADL videos often involve actions with \textbf{similar appearances}, \textbf{subtle motions}, and may be captured from \textbf{multiple viewpoints}. These characteristics limit the ``\textit{generalizability}'' of video models trained on web videos to those containing ADL.
Consequently, this \textit{distributional shift} in video representations causes the video models to struggle in recognizing action categories not encountered during training as illustrated in Figure~\ref{fig:teaser}. As a result, these models lack robustness as zero-shot learners, a crucial characteristic necessary for monitoring systems to detect anomalous human gestures indicative of cognitive decline. This motivates us to develop vision-language models (VLMs) capable of generalizing to unseen ADL scenarios. While the aforementioned \textit{distribution shift} can be mitigated by training VLMs~\cite{vificlip, XCLIP, FROSTER} on ADL videos~\cite{NTU_RGB+D, ntu120}, these models still struggle to effectively address the inherent challenges in ADL due to the lack of specialized operations within VLMs designed to tackle them.

Instead of relying on specialized architectural changes, many works have explored the use of additional modalities beyond vision for action recognition. For instance, optical flow has been effective in web video datasets~\cite{kinetics, ucf, kuehne2011hmdb}, where actions are characterized by strong motion
~\cite{mars, twostream}. 
However, it is less effective on ADL datasets~\cite{NTU_RGB+D, ntu120, smarthome}, where motion cues are more subtle~\cite{vpn++}. Conversely, 3D skeletons have been established as a strong modality, being viewpoint-agnostic and providing crucial information for modeling ADL
~\cite{stgcn, 2sagcn2019cvpr,zhou2022hyperformer}. 
Similar to VLMs, skeleton-language representations have been explored in~\cite{synse, smie, star}. 
Skeleton action recognition models, while excelling at capturing motion cues and human motion, do not encode the appearance information crucial for distinguishing nuanced human-object interactions. For instance, actions like \textit{Drinking from Cup} versus \textit{Drinking from Bottle} rely on visual appearance to differentiate objects, which motion alone cannot provide. Thus, integrating appearance information can enhance model’s ability to accurately classify ADLs. This is demonstrated in~\cite{vpn++}, where appearance-based methods outperform approaches that rely solely on skeleton information. 
Additionally, the combination of RGB and skeletons has been investigated in
~\cite{das2020vpn, PoseC3D_CVPR22, vpn++, pivit}, 
but these approaches are not zero-shot learners.
Therefore, in this paper, we pose the critical question: \textit{Can we introduce 3D skeletons into the vision-language embedding space to enable effective zero-shot learning for ADL?}

One naive solution is to learn a common embedding space for videos, skeletons, and language. However, the limited availability of large-scale synchronized data renders this approach ineffective. Another approach involves aligning skeleton embeddings with learned image embeddings~\cite{imagebind} or language embeddings~\cite{LanguageBind}. However, skeletons require neural networks capable of modeling the implicit spatial configuration of human body kinematics, which differ from traditional ViT encoders. This discrepancy limits their alignment with image or language embeddings, as effective projection within the CLIP embedding space typically requires homogeneous modality-specific encoders.
Therefore, we introduce a series of \textbf{SK}eleton \textbf{I}nduced models, referred to as \textbf{SKI} models, that effectively integrate skeletons within vision-language embeddings, enabling them to understand ADL. Our SKI models are classified as \textbf{SKI-VLMs} when skeletons are induced within VLMs, and \textbf{SKI-LVLMs} when skeletons are integrated into large vision-language models (LVLMs). The core of SKI models is based on a skeleton-language model, termed \textbf{SkeletonCLIP}, which successfully incorporates 3D human skeleton knowledge into the vision-language embeddings. This integration guides the vision encoders to focus on human joints for learning action representations (Fig.~\ref{fig:teaser}).
 
The training of SKI-VLMs involves first aligning modality-specific features (videos or skeletons) with language, followed by performing knowledge distillation (KD) in a language-contextualized space. This KD process, termed SkeletonCLIP Distillation (SCD), involves the collaborative training of SkeletonCLIP and the VLM while conducting KD. 
Similarly, the training of SKI-LVLMs involves integrating 3D skeleton features extracted from the SkeletonCLIP encoder to enhance the semantic reasoning and generation capabilities of LVLMs. SKI-LVLMs can be trained by learning an additional skeleton projector, which maps skeleton-language tokens into the embedding space of the large language models (LLMs) within the LVLMs. 
During inference of SKI models, SkeletonCLIP and its components can be discarded, resulting in a skeleton-augmented VLM or LVLM capable of inferring human skeleton knowledge, which is crucial for understanding ADL.
We conduct extensive experimental evaluation of SKI-VLM using three VideoCLIP dual encoders to assess its effectiveness in zero-shot action recognition. Additionally, we evaluate the performance of SKI-LVLM on dense video caption generation.

\noindent To summarize our contributions:
\begin{enumerate}
    \item We introduce Skeleton-Induced VLM (SKI-VLM), an effective approach for integrating 3D skeleton information into the VLM space using SkeletonCLIP Distillation (SCD). The resulting SKI-VLM addresses the challenges of ADL by focusing on human key points while modeling action representations.
    \item We present SKI-LVLM, which incorporates language-grounded 3D skeletal features as an additional modality in LVLMs, enhancing their video understanding capabilities.
    \item We demonstrate the superior performance of SKI-VLM in zero-shot action recognition on the largest RGB+D datasets: NTU60~\cite{NTU_RGB+D} and NTU120~\cite{ntu120}. We also evaluate SKI-LVLM's ability to generate text descriptions on the Charades dataset~\cite{charades}, which comprises dense captions. SKI-LVLM outperforms the baseline, highlighting the importance of incorporating skeleton information into LVLMs.
\end{enumerate}

In practice, SKI models can be implemented with any skeleton-language models and VLMs. To the best of our knowledge, this is the first work that attempts to enhance vision-language embeddings by incorporating skeleton information for video representation learning.

\section{Related Work}
\noindent
\textbf{Multi-modal Knowledge Distillation.}
In standard knowledge distillation~\cite{knowledge_distillation_hinton2015}, the knowledge of large-scale models (teachers) is transferred to smaller models (students), enabling students to replicate the teacher’s predictions or feature representations and replace them during inference. Similarly, multi-modal knowledge distillation transfers the knowledge of a teacher model to a student model operating on a different modality. This approach is effective for tasks like action recognition, where complementary modalities exist that are not available or infeasible to compute at inference time (e.g., optical flow or human skeleton). In action recognition, the student modality is typically RGB, and the teacher modality is optical flow~\cite{gupta_crossmodaldistillation_cvpr16, mars} or audio~\cite{soundnet}. However, these modalities are limited in modeling human motion and generalizing across varying viewpoints, limiting their adoption for ADL~\cite{das2020vpn}. Human skeletons, by contrast, have emerged as the dominant modality to combine with RGB for ADL tasks~\cite{das2020vpn, vpn++, pivit}, as they effectively model human motion and generalize across viewpoints. While effective for understanding ADL, these previous approaches are not applicable to zero-shot tasks. In contrast, our proposed SKI-VLM and SKI-LVLM overcome this limitation by incorporating both modalities with language, enabling zero-shot tasks.\\
\textbf{Multi-modal VLMs for Action Recognition.}
Many approaches propose to extend the image-based CLIP to the video domain through fine-tuning CLIP to handle the temporal dimension of video data. Partially fine-tuned approaches
~\cite{yang2023_AIM_ICLR, XCLIP, st_adapter}
perform training with additional trainable temporal modules but leave CLIP's parameters frozen, while fully fine-tuned approaches~\cite{ActionCLIP, vificlip, FROSTER} perform a simple temporal pooling but update the CLIP parameters during training. These approaches only process RGB and ignore the rich multi-modal nature of videos. In response, some works attempt to incorporate additional modalities such as optical flow~\cite{qian2022_opticalflow_and_audioclip} and audio~\cite{Audioclip, clip4vla, Wav2CLIP} into the CLIP embedding space. While these approaches all aim to introduce new modalities into CLIP, their methodologies vary. For example, \cite{Audioclip} trains audio, visual, and language encoders using a tri-modal contrastive loss, while \cite{Wav2CLIP} contrastively aligns an audio encoder with a frozen CLIP visual encoder. Different from these works, we introduce the skeleton modality into the CLIP space to better address the challenges of ADL. Additionally we find, and experimentally validate, that previous alignment strategies are sub-optimal when considering the skeleton modality.

\textbf{Zero-Shot Skeleton Action Recognition.}
Zero-shot skeleton-based models aim to enable action classification of unseen classes using only human skeleton sequences. SynSE~\cite{synse} introduces a syntactically guided approach, using part-of-speech tags to enhance the alignment between skeleton and language spaces. SMIE~\cite{smie} employs mutual information estimation and maximization to globally align the two spaces. CrossGLG~\cite{crossglg} uses LLM descriptions of actions and cross-attention to guide a skeleton encoder during training, but only uses the skeleton encoder at inference. Closest to our work is STAR~\cite{star}, which aligns a Shift-GCN~\cite{cheng2020shiftgcn} skeleton encoder with a pre-trained transformer-based text encoder~\cite{CLIP} for zero-shot skeleton action recognition. STAR differs from our work in that it does not incorporate the RGB modality or investigate strategies to enhance the representations of VLMs and LVLMs.


\textbf{Multi-modal Large Language Models.}
Large language models, such as ChatGPT~\cite{chatgpt} and LLaMA~\cite{llama}, have exhibited remarkable capabilities in language understanding and generation tasks. These capabilities have inspired many works to extend LLMs to incorporate additional modalities, such as video~\cite{videollama, videollava, videochatgpt}, audio~\cite{zhang2023_speechgpt}, and human pose~\cite{feng2024_chatpose}, typically through instruction tuning or the addition of modality-specific encoders with projection layers. Extending beyond this, other approaches show the possibility to incorporate a wide range of modalities into the LLM space
~\cite{lu2022_unifiedio, su2023_pandagpt}. Unique 
from these approaches, our SKI-LVLM targets ADL and aims to train using multiple modalities (vision, language and skeleton), but only use vision and language during inference.
\section{Proposed Method}

In this section, we present SkeletonCLIP, a specialized skeleton-language model designed to align 3D skeleton representations of human actions with their corresponding language representations. SkeletonCLIP is a pivotal element of our approach and can be substituted with any skeleton-text model~\cite{smie, synse, star} to enhance the VLM and LVLM embedding space. This is achieved by leveraging its capacity to capture fine-grained motion details and multiple viewpoints effectively. 
To integrate the crucial skeleton representation into VLMs, we first propose the \textbf{Skeleton-Induced VLM} (\textbf{SKI-VLM}), which seamlessly incorporates SkeletonCLIP features through \textbf{SkeletonCLIP Distillation} (\textbf{SCD}).  
Furthermore, we introduce the \textbf{Skeleton-Induced LVLM} (\textbf{SKI-LVLM}), wherein SkeletonCLIP features are included as an additional modality within LVLMs. This integration enhances the capability of these models to interpret nuanced and complex actions in video data.


\begin{figure*}[h!]
  \centering
    \includegraphics[height=8.2cm, width=15cm]{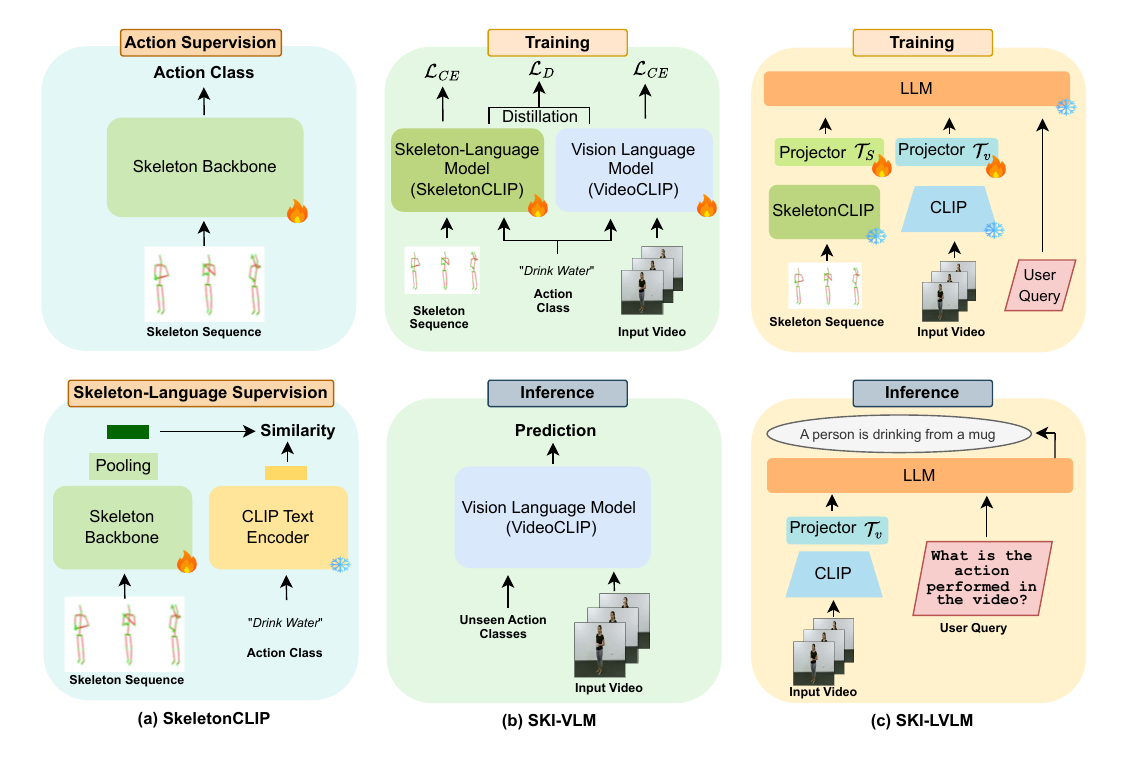}
  \caption{
    \textbf{(a) SkeletonCLIP}: Utilizes a pretrained Skeleton Backbone, aligned with action class labels from the frozen CLIP Text Encoder during Skeleton-Language Supervision.
    \textbf{(b) SKI-VLM}: Engages in online distillation between SkeletonCLIP and a Vision-Language Model (VLM), both trainable. For inference, the VLM alone performs zero-shot action recognition on unseen classes.
    \textbf{(c) SKI-LVLM}: Projects SkeletonCLIP features into LLM space along with video features. Only the projection layers are trainable, while SkeletonCLIP, CLIP encoder, and LLM are frozen. Inference uses the CLIP vision encoder to extract video features, which, together with the user query, are input to the LLM to generate a response based on the video content.
    \begin{minipage}{0.6\textwidth}
        \includegraphics[width=0.05\textwidth]{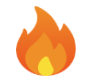}
        \raisebox{0.7ex}{\makebox[1.5cm][l]{Trainable}}
        \includegraphics[width=0.05\textwidth]{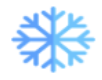}
        \raisebox{0.7ex}{\makebox[1cm][l]{Frozen}}
    \end{minipage}}
\label{fig:architecture}
\end{figure*}

\subsection{SkeletonCLIP}
\label{subsec:SkeletonCLIP}

SkeletonCLIP is a dual encoder Skeleton-Language model that jointly learns 3D Skeleton and language representations within a common semantic space as illustrated in Figure~\ref{fig:architecture} (left). Unlike VLMs~\cite{CLIP, vificlip}, SkeletonCLIP faces constraints in utilizing extensive skeleton-text data due to the scarcity of synchronized 3D skeleton sequences and corresponding textual data. It comprises two primary components: a Skeleton Encoder $g_s$ and a Text Encoder $g_t$. We chose Hyperformer \cite{zhou2022hyperformer} as the skeleton encoder due to its ability to learn the kinematic dependencies between human joints. This model has shown notable efficacy in Skeleton-based Action Recognition, making it an ideal choice as a skeleton encoder in our approach. Conversely, the text encoder leverages the CLIP text encoder, capitalizing on CLIP's discriminative representation capabilities to counterbalance the limited availability of large-scale skeleton sequence data.
SkeletonCLIP is a plug-and-play module that can be integrated with any VLM, such as XCLIP, ViFiCLIP, or FROSTER, facilitating skeleton-guided visual representation learning.   

SkeletonCLIP processes input skeleton sequences $S_i \in \mathbb{R}^{T_s \times 3 \times J}$, where each frame within the sequence $T_s$ comprises $J$ 3D skeleton joints, alongside a text prompt $t_j$ representing the action class label. Initially, the skeleton encoder $g_s$ is pretrained for the task of skeleton action recognition. This pretraining enables SkeletonCLIP to learn a joint embedding with language using an already pretrained CLIP~\cite{CLIP} text encoder $g_t$. 

The skeleton encoder yields a skeleton representation $z_i^s =  \frac{1}{T_s}\sum g_s(S_i)$ for an input sequence $S_i$ from its penultimate layer, whereas the text encoder $g_t$ yields a feature $z_j^t = g_t(t_j)$ for each action class text prompt $t_j$. To learn a joint skeleton-language embedding, the pretrained skeleton and text encoders are trained jointly while keeping the text encoder frozen.
This joint embedding is optimized by maximizing the cosine similarity between skeleton and text embeddings, quantified by the cross-entropy loss:
\begin{equation}
\mathcal{L}_{CE}(z_i^s, z_i^t) = - \sum_i \log  \frac{\exp(sim(z_i^s, z^t_i)/\tau)}{\sum_j\exp(sim(z_j^s, z_j^t)/\tau)}
\end{equation}
where $\tau$ is a temperature parameter and $sim(x, y)$ denotes the cosine similarity between $x$ and $y$.
This joint training enforces the skeleton encoder to align with the language embeddings, enabling it to perform zero-shot skeleton action recognition. This language grounding is essential for incorporating skeleton features into VLMs and LVLMs, which is detailed next.

\subsection{Skeleton Induced VLM (SKI-VLM)}
Consider a video $V_i \in \mathbb{R}^{T_v \times 3\times H\times W}$ comprising $T_v$ frames at a spatial resolution of $H \times W$  and a text prompt $t_j$ encapsulating the action categories within a predefined template. To obtain their representations within the VLM embedding space, the input videos and text prompts are processed through the CLIP~\cite{CLIP} image encoder $f_v$ and text $f_t$ encoder respectively as
\begin{equation}
    z^v_i = \frac{1}{T_v}\sum f_v(V_i);  \hspace{0.4in}  z^t_j = f_t(t_j)  
\end{equation}
The video-level representation $z^v_i$, derived from the CLIP image encoder, is obtained by averaging the feature representations across frames. Typically, the joint embedding ($z_i^v \cdot z_j^t$) is trained using various finetuning techniques for learning discriminative video representation. These include fully finetuning~\cite{vificlip} the CLIP embeddings using $\mathcal{L}_{CE}(z_i^v, z_j^t)$, partially finetuning~\cite{XCLIP} using adaptors, or fully finetuning in addition to knowledge distillation from CLIP~\cite{FROSTER}. The trained dual encoder model is referred to as a VideoCLIP model, capable of performing zero-shot action recognition in RGB videos.

To integrate 3D skeleton information into the joint video-text embeddings ($z_v \cdot z_t$), we leverage an online knowledge distillation strategy, \textit{SkeletonCLIP Distillation} (SCD). It is performed across the Skeleton-Language (SkeletonCLIP) and Vision-Language (VideoCLIP) models, both of which are pretrained on the training distribution. This pretraining ensures the alignment of the modality-specific features ($z_i^v$ or $z_i^s$) with the textual features ($z_i^t$). Since the textual features differ across modalities (videos and skeletons), hereon we will denote the video-based and skeleton-based textual representations as $z_i^{tv}$ and $z_i^{ts}$, respectively.

\noindent
\textbf{SkeletonCLIP Distillation (SCD).}
For SCD, the model with two dual encoders takes as input video sequences $V_i$, corresponding 3D skeleton sequences $S_i$, and text prompts $t_j$ describing action labels, as illustrated in Figure~\ref{fig:architecture} (middle). The VideoCLIP processes the video sequences and text prompts using the encoders $f_v$ and $f_t$, respectively. Simultaneously, the SkeletonCLIP processes the 3D skeleton sequences and text prompts using the encoders $g_s$ and $g_t$, respectively. 
SCD is then applied to integrate visual and skeleton information within a language-contextualized space. This space is constructed by minimizing the similarity between modality-specific features ($z_i^v$ or $z_i^s$) and text features ($z_i^{tv}$ or $z_i^{ts}$). The resulting language-contextualized features in the visual and skeleton domains, $F_{LV}$ and $F_{LS}$, are defined as $F_{LV} = z_i^v \cdot (z_j^{tv})^T$ and $F_{LS} = z_i^s \cdot (z_j^{ts})^T$.

We adopt an online knowledge distillation approach where both the SkeletonCLIP and VideoCLIP models are trainable. This distillation, conducted within the language-contextualized space, aims to minimize the similarity between the respective modalities and the text features, thereby enabling VideoCLIP to infer skeleton-text correlations when applicable. To ensure the relevance of 3D skeleton information for action recognition, we jointly train the SCD with action recognition losses.
The total loss ($\mathcal{L}$) during this training phase is represented as the sum of the cross-entropy losses for video-text and skeleton-text embeddings, combined with a distillation loss $\mathcal{L}_{D}$ (mean squared error):
\begin{equation}
\mathcal{L} = \mathcal{L}_{CE}(z_i^v, z_i^{tv}) + \mathcal{L}_{CE}(z_i^s, z_i^{ts}) + \alpha \mathcal{L}_{D}(F_{LV}, F_{LS})
\end{equation}
where $\alpha$ is the weight allocated to the distillation loss to balance it with the cross-entropy losses. Through SCD, SkeletonCLIP enriches VideoCLIP with temporal information, resulting in the Skeleton-Induced VideoCLIP (SKI-VLM). 

During \textbf{inference}, only the VideoCLIP (SKI-VideoCLIP) is utilized, eliminating the need for skeleton data and avoiding additional computational overhead. We implement a teacher-student framework to learn the SKI-VLM representation through SCD, which can be integrated with various student VLMs~\cite{vificlip, XCLIP, FROSTER}.


\subsection{Skeleton Induced LVLM (SKI-LVLM)}
3D skeleton features can also be employed to guide visual features, thereby enhancing the semantic reasoning and generation capabilities of Large Vision Language Models (LVLMs). These enhancements are expected to improve the quality of descriptions generated by LVLMs, as they can focus on human joints and their motion while generating textual content. To integrate skeleton information within LVLMs, the features $z_i^s$ for an input skeleton sequence are first extracted from SkeletonCLIP. Given the effectiveness of SCD, where collaborative learning of modality-specific dual encoders proved highly effective, we adopt a similar design approach for integrating skeletons within LVLMs.

In this approach, both visual ($z_i^v$) and skeleton ($z_i^s$) features extracted from the RGB+skeleton input sequence are fed into their respective projection layers ($\mathcal{T}_v$ and $\mathcal{T}_s$), which are then input to the LLM:
\begin{equation}
    Q_v = \mathcal{T}_v(z_i^v) \in \mathbb{R}^{F_{v} \times K}, \quad
    Q_s = \mathcal{T}_s(z_i^s) \in \mathbb{R}^{F_{s} \times K}
\end{equation}
where $F_{v}$ and $F_{s}$ represent the dimensionality of the video and skeleton features, respectively, and $K$ is the shared dimensionality to which these features are projected. Similar to the visual projector $\mathcal{T}_v$, the skeleton projector $\mathcal{T}_s$ learns the mapping of skeleton tokens from the skeleton-language space to the input space of the LLM. Notably, the visual features here are extracted from the CLIP encoder without finetuning on the training distribution, to maintain the generalizability of the RGB information fed to the LLM.

Thus, the input to the LLM consists of $Q_v$, $Q_s$, and the tokenized text query $Q_t \in \mathbb{R}^{F_{t} \times K}$ (where $F_{t}$ is the dimension of the text features), arranged in the following template: $ [   \textbf{USER:} \hspace{0.05in} \langle Q_t \rangle \hspace{0.05in} \langle Q_v \rangle \hspace{0.05in} \langle Q_s \rangle, \hspace{0.05in} \textbf{Assistant:}]$.
SKI-LVLM is trained collaboratively on video-skeleton-text triplets using the autoregressive training objective from~\cite{videochatgpt}. The weights of the encoders and the LLM are frozen, and only the projection layers are trained, as illustrated in Figure~\ref{fig:architecture} (right). During \textbf{inference}, only the visual input is utilized, eliminating the need for skeleton data and its projector $\mathcal{T}_s$, making it practical for real-world applications. 

\section{Experimental Results}
\label{sec:experiments}

\noindent \textbf{Datasets.} For \textit{zero-shot} (ZS) \textit{action recognition}, we evaluate our SKI-VLMs on the large-scale NTU-RGB+D-60 (NTU60) \cite{NTU_RGB+D} and NTU-RGB+D-120 (NTU120) \cite{ntu120} datasets. NTU120 contains approximately 114K video-pose pairs across 120 action classes, while NTU60 includes around 57K pairs for 60 classes. We adopt the evaluation splits from \cite{synse}, using 55/5 and 48/12 splits for NTU60, and 110/10 and 96/24 splits for NTU120. 

For \textit{dense video captioning}, we train our SKI-LVLM on NTU120 video-instruction pairs and evaluate on the Charades dataset~\cite{charades} following \cite{llavidal2024}. NTU120 video-instruction pairs are generated by captioning single frames (mid-video) with COGVLM~\cite{cogvlm} and creating 100K question-answer pairs using GPT 3.5 turbo. Details on prompts are in the Appendix~\ref{sec:lvlm_data}. We will release these video-instruction pairs to the community.
We evaluate video captioning performance using Llama 3.1~\cite{meta2024_llama3herdmodels} on the five VideoChatGPT~\cite{videochatgpt} metrics: \textit{Correctness of Information, Detail Orientation, Contextual Understanding, Temporal Understanding} and \textit{Consistency}.

\noindent \textbf{Implementation Details.} 
For \textbf{SKI-VLMs}, we use XCLIP \cite{XCLIP}, ViFiCLIP \cite{vificlip}, and FROSTER \cite{FROSTER} as the student VLMs (VideoCLIP). SkeletonCLIP is trained on seen classes of NTU60 (55, 48) and NTU120 (110, 96) for action recognition. The skeleton encoder is pretrained for 140 epochs, followed by alignment with a CLIP Text encoder for 100 epochs. SCD requires 2 epochs. We use a learning rate of $2.25 \times 10^{-5}$ with cosine decay and set $\alpha$ to 0.01 for NTU60 and 10.0 for NTU120. SKI-XCLIP, SKI-ViFiCLIP, and SKI-FROSTER denote SKI-VLM with respective backbones.
For \textbf{SKI-LVLM}, embedding dimensions are $F_v=1024$, $F_s=216$, $K=4096$. The input to the projection layers ($\mathcal{T}_v$ \& $\mathcal{T}_s$) preceding the LLM are 356 visual and 256 skeleton tokens. We train SKI-LVLM and its baselines for 3 epochs with batch size 32, learning rate $2e^{-5}$ on 8 A6000 48GB GPUs. 

\begin{table}[h]
\centering
\scriptsize 
\setlength{\tabcolsep}{3pt} 
\renewcommand{\arraystretch}{0.75} 
\begin{tabular}{lccc|cc}
\toprule
\textbf{Method} & \multicolumn{3}{c|}{\textbf{Modality}} & \textbf{NTU60} & \textbf{NTU120} \\
                & \textbf{V} & \textbf{T} & \textbf{S} & \textbf{48/12} & \textbf{110/10} \\
\midrule
Tri-modal Align.~\cite{Audioclip} & \xmark & $\checkmark$ & $\checkmark$ & 8.3 & 11.9 \\
Cross-projection Align.~\cite{Wav2CLIP} & \xmark & $\checkmark$ & $\checkmark$ & 8.4 & 25.5 \\
\midrule
XCLIP~\cite{XCLIP}         & $\checkmark$ & $\checkmark$ & \xmark & 38.9 & 57.8 \\
FROSTER~\cite{FROSTER}         & $\checkmark$ & $\checkmark$ & \xmark & 43.9 & 65.2 \\
ViFiCLIP~\cite{vificlip}        & $\checkmark$ & $\checkmark$ & \xmark & 48.2 & 70.0 \\
\midrule
SkeletonCLIP & \xmark & $\checkmark$ & $\checkmark$ & 35.1 & 63.0 \\
ViFiCLIP + SkeletonCLIP & $\checkmark$ & $\checkmark$ & $\checkmark$ & 17.5 & 48.8 \\
\midrule
\rowcolor{gray!25}
\textbf{SKI-XCLIP} & $\checkmark$ & $\checkmark$ & $\circ$ & 42.2 & 66.1 \\
\rowcolor{gray!25} \textbf{SKI-FROSTER} & $\checkmark$ & $\checkmark$ & $\circ$ & 44.4 & 68.5 \\
\rowcolor{gray!25} \textbf{SKI-ViFiCLIP} & $\checkmark$ & $\checkmark$ & $\circ$ & \textbf{52.0} & \textbf{77.5} \\
\bottomrule
\end{tabular}
\caption{Performance comparison of methods for aligning skeleton features with video-text embeddings on NTU60 and NTU120 datasets, demonstrating the effectiveness of integrating language-contextualized 3D skeleton features into the CLIP embedding space. \textit{V}, \textit{S} and \textit{T} denote the video, skeleton, and text modalities respectively. \footnotesize \textit{($\circ$ indicates that skeleton features were used only during training)}}
\label{tab:vpt_emd}
\end{table}

\subsection{ZS Action Recognition using SKI-VLM}
For zero-shot action recognition, we evaluate models on unseen action classes within each split. We first demonstrate the challenge of learning a common embedding space for video, text, and skeletons. We then justify the superiority of VideoCLIP architectures over SkeletonCLIP and highlight the robustness of SKI-VLMs. Finally, we present state-of-the-art zero-shot action recognition results on NTU datasets.

\noindent \textbf{How can we introduce 3D skeleton features in the CLIP embedding space?}
In the audio domain, methods such as AudioCLIP~\cite{Audioclip} and Wav2CLIP~\cite{Wav2CLIP} employ contrastive learning and cross-projection techniques to align audio with the CLIP embedding space. Inspired by these approaches, we explore aligning skeleton features with the video-text embedding space. However, as shown in Table~\ref{tab:vpt_emd}, neither tri-modal alignment (implemented following AudioCLIP) nor cross-projection alignment (implemented following Wav2CLIP) effectively aligns skeleton and text representations. This limitation stems from the lack of large-scale symmetric video-skeleton-text datasets necessary for effective contrastive learning. The results of SKI-VLMs highlights the effectiveness of integrating language-contextualized 3D skeleton features into the vision-language embedding space, enhancing the learning of generalized action representations.\\
\noindent \textbf{SkeletonCLIP vs VideoCLIP.}
SkeletonCLIP grounds skeleton features within a language-contextualized space. As shown in Table~\ref{tab:vpt_emd}, SkeletonCLIP achieves performance comparable to VideoCLIP models like XCLIP, FROSTER, and ViFiCLIP. To further enhance zero-shot recognition performance, we combined the strengths of VideoCLIP and SkeletonCLIP by directly fusing language-contextualized skeleton features with VLM features. However, this approach led to poorer action recognition performance, likely due to conflicting gradients from the different modalities. Thus, VLMs that incorporate only video and text modalities demonstrate more promising performance in zero-shot action recognition compared to skeleton-only models.\\
\noindent \textbf{Robustness of SKI-VLMs.}
Our SKI-VLMs can be implemented using any existing dual encoders. In Table~\ref{tab:vpt_emd}, we demonstrate the implementation of SKI-VLMs with XCLIP, ViFiCLIP, and FROSTER, where they outperform their respective baselines by up to \textcolor{green}{+14.3\%}, \textcolor{green}{+10.7\%}, and \textcolor{green}{+5.1\%}. This highlights the robustness of our SKI-VLMs when integrated with any student VLM.

\begin{table}[h]
\centering
\scriptsize 
\setlength{\tabcolsep}{2pt} 
\renewcommand{\arraystretch}{0.75} 
\begin{tabular}{>{\arraybackslash}m{3.3cm}|>{\centering\arraybackslash}m{0.3cm}>{\centering\arraybackslash}m{0.3cm}>{\centering\arraybackslash}m{0.3cm}|>{\centering\arraybackslash}m{0.7cm}>{\centering\arraybackslash}m{0.7cm}|>{\centering\arraybackslash}m{0.7cm}>{\centering\arraybackslash}m{0.7cm}}
\toprule
\textbf{Method} & \multicolumn{3}{c|}{\textbf{Modality}} & \multicolumn{2}{c|}{\textbf{NTU60}} & \multicolumn{2}{c}{\textbf{NTU120}} \\
                & \textbf{V} & \textbf{T} & \textbf{S} & \textbf{55/5} & \textbf{48/12} & \textbf{110/10} & \textbf{96/24} \\
\midrule
SynSE~\cite{synse}     & \xmark       & $\checkmark$  & $\checkmark$  & 75.8 & 33.3 & 62.7 & 38.7 \\
SMIE~\cite{smie}            & \xmark       & $\checkmark$      & $\checkmark$ & 78.0    & 40.2    & 65.7    & 45.3    \\
STAR~\cite{star}            & \xmark       & $\checkmark$      & $\checkmark$ & \underline{81.4}    & 45.1    & 63.3    & 44.3    \\
CLIP~\cite{CLIP}            & $\checkmark$   & $\checkmark$      & \xmark      & 54.5    & 20.2 & 35.0    & 15.2    \\
XCLIP~\cite{XCLIP}           & $\checkmark$   & $\checkmark$      & \xmark      & 76.0    & 38.9    & 57.8    & 49.7    \\
ViFiCLIP~\cite{vificlip}        & $\checkmark$   & $\checkmark$  & \xmark  & 79.9 & \underline{48.2} & \underline{70.0} & \underline{56.6} \\
FROSTER~\cite{FROSTER}         & $\checkmark$   & $\checkmark$      & \xmark      & 79.1    & 43.9    & 65.2    & 34.0    \\
\midrule 
\rowcolor{gray!25} \textbf{SKI-ViFiCLIP}   & $\checkmark$   & $\checkmark$  & $\circ$     & \textbf{82.2} & \textbf{52.0} & \textbf{77.5} & \textbf{59.3} \\
\bottomrule
\end{tabular}
\caption{Comparison of Zero-Shot Action Recognition Accuracy on NTU60 and NTU120 datasets. \textit{V}, \textit{S} and \textit{T} denote the video, skeleton, and text modalities respectively. \footnotesize \textit{($\circ$ indicates that skeleton features were used only during training)}}
\label{tab:zsl} 
\end{table}

\noindent \textbf{State-of-the-art comparison.}
In Table~\ref{tab:zsl}, we benchmark SKI-VLMs against leading skeleton-text and video-text zero-shot action recognition models. The inferior performance of skeleton-text models like SynSE \cite{synse} and SMIE \cite{smie} can be attributed to their lack of appearance information. Furthermore, adapting these models to different data distributions is challenging due to variations in skeleton configurations arising from disparate depth sensors~\cite{MicrosoftKinuct} or skeleton extraction techniques~\cite{lcrnet_new}. 
We also note that FROSTER~\cite{FROSTER}, despite outperforming ViFiCLIP~\cite{vificlip} and XCLIP~\cite{XCLIP} in web video datasets such as Kinetics~\cite{kinetics}, UCF-101~\cite{ucf}, and HMDB~\cite{kuehne2011hmdb}, falls short in ADL datasets, showing the unique challenges posed by ADL compared to web videos. FROSTER’s reliance on image-based CLIP knowledge transfer is less effective on NTU due to the lack of scene-specific contextual information in NTU.

SKI-ViFiCLIP surpasses all other zero-shot action recognition models. Notably, applying SCD to a video-based CLIP model (ViFiCLIP) significantly enhances its zero-shot action recognition performance on all NTU60 and NTU120 splits by up to \textcolor{green}{+5.4\%} and \textcolor{green}{+7.8\%}, respectively. This demonstrates the enhanced "generalizability" of SKI-VLM when incorporating skeletons into VLMs.

\begin{table}
\centering
\scriptsize 
\setlength{\tabcolsep}{2pt} 
\renewcommand{\arraystretch}{0.75} 
\begin{tabular}{lccaa}
\toprule
\textbf{Metric} & \textbf{VCGPT} & \textbf{NTU-VCGPT} & \textbf{SK-VCGPT} & \textbf{SKI-LVLM} \\
\midrule
Correctness & 24.8 & 25.5 & 40.6 & 40.6 \\
Detail Orientation & 43.2 & 21.6 & 55.0 & 54.8 \\
Contextual Understanding & 33.8 & 29.0 & 52.4 & 52.4 \\
Temporal Understanding & 20.0 & 16.9 & 30.6 & 31.6 \\
Consistency & 31.6 & 34.1 & 38.4 & 38.2 \\
\midrule
\textbf{Average} & 30.7 & 25.4 & 43.4 & \textbf{43.5} \\
\bottomrule
\end{tabular}
\caption{Dense Caption Generation Performance of SkeletonCLIP induced LVLM on Charades Dataset (VCGPT = VideoChatGPT)}
\label{tab:skllm}
\end{table}

\subsection{Dense Video Captioning using SKI-LVLM}
In Table~\ref{tab:skllm}, we assess the video captioning performance of SKI-LVLM using the five metrics proposed in~\cite{videochatgpt}, scaled to a range of 1-100 following \cite{llavidal2024}. 
SKI-LVLM is benchmarked against VideoChatGPT (VCGPT) and NTU trained VideoChatGPT (NTU-VCGPT). VCGPT is trained on a large-scale Instruction Tuning dataset derived from ActivityNet~\cite{caba2015activitynet}, while NTU-VCGPT denotes VideoChatGPT trained exclusively on video-instruction pairs from trimmed NTU120 videos (see Ablation \ref{sec:lvlm_data} for details). Additionally, we include a variant of SKI-LVLM, denoted SK-VCGPT, which is trained on NTU120 instruction pairs and corresponding 3D skeleton features obtained from a pre-trained skeleton backbone (Hyperformer). This allows us to investigate the value of language-contextualized skeleton features in SKI-LVLM, as SK-VCGPT incorporates raw 3D skeleton features without language contextualization.

Surprisingly, we find that SK-VCGPT performs comparably with SKI-LVLM, contrasting with our findings in traditional VLMs where language contextualization is essential. This suggests that feature integration mechanisms may differ in LVLMs, indicating that language contextualization in LVLMs plays a less critical role. However, we observe that the introduction of 3D skeletons greatly improves the performance of the LVLMs, highlighting the importance of integrating skeleton features into LVLM training.

\begin{table}
\centering
\scriptsize 
\renewcommand{\arraystretch}{0.75} 
\begin{tabular}{lccc}
\toprule
\textbf{Method} & \textbf{NTU60} & \textbf{NTU120} & \textbf{Harmonic} \\
& \textbf{48/12}  & \textbf{110/10} & \textbf{Mean} \\
\midrule
Feature-level KD \textit{w/o Projection} & 52.9 & 65.3 & 58.4 \\
Feature-level KD \textit{w Projection} & 53.2 & 64.7 & 58.4 \\
Offline KD & \textbf{53.3} & 70.8 & 60.8 \\
\rowcolor{gray!25}\textbf{Online KD (ours)} & 52.0 & \textbf{77.5} & \textbf{62.2} \\
\bottomrule
\end{tabular}
\caption{Performance on different Distillation techniques}
\label{tab:kd_abltn}
\end{table}

\subsection{Ablation Study}
All our ablation studies have been conducted with ViFiCLIP as the dual encoder (VideoCLIP).

\noindent \textbf{Variants of Knowledge Distillation (KD).}
KD between SkeletonCLIP and VideoCLIP can be implemented using different strategies. In Table~\ref{tab:kd_abltn}, we compare SCD with feature-level distillation and offline KD. Feature-level distillation, used in video-skeleton action recognition to enhance RGB-based encoders~\cite{vpn++, pivit}, is sub-optimal for large-scale datasets in SKI-VLMs. Feature space mismatches between the teacher (SkeletonCLIP) and student (VideoCLIP) are magnified in large-scale settings, leading to inefficiencies in knowledge transfer. Online distillation at the logit level, directly aligning network outputs during training, offers a more robust solution for handling large-scale data. Additionally, using a projection layer to map features into a common space for distillation, as explored in~\cite{FROSTER}, improves performance on NTU60 but doesn’t consistently boost accuracy across larger datasets. Unlike online KD in SCD, where both networks are trainable, offline KD freezes the teacher and trains only the student. Our findings suggest that while offline and feature-level KD perform better in low-data settings, online KD consistently outperforms them in large-scale datasets. This highlights the benefit of collaborative training between SkeletonCLIP and VideoCLIP, allowing the latter to acquire meaningful representations and achieve higher accuracy.

\begin{table}[]
\centering
\scriptsize 
\renewcommand{\arraystretch}{0.75} 
\begin{tabular}{lccc|cc}
\toprule
\textbf{Loss Function} & \textbf{NTU60} & \textbf{NTU120} \\
& 48/12 & 110/10 \\
\midrule
Contrastive Loss & 46.9 & 70.0 \\
KL Divergence Loss & 48.5 & 73.1 \\
\rowcolor{gray!25} \textbf{Mean Squared Error (MSE) Loss} & \textbf{52.0} & \textbf{77.5} \\
\bottomrule
\end{tabular}
\caption{Effect of choice of distillation loss functions}
\label{tab:con}
\end{table}

\noindent
\textbf{Loss Configuration.} In Table~\ref{tab:con}, we perform an ablation study on the choice of distillation loss for the implementation of SCD. We find that employing MSE for SCD results in notably improved performance compared to using Contrastive Loss or KL Divergence Loss.

\section{Conclusion}
In this paper, we introduced Skeleton-Induced (SKI) models, including SKI-VLMs and SKI-LVLMs, to integrate skeleton information into vision-language embeddings. This enables the models to focus on human keypoints for better action modeling, crucial for learning discriminative video representations in ADL. Our experiments show that SKI-VLMs achieve significant gains in zero-shot action recognition without needing skeletons during inference, while SKI-LVLM improves LVLMs’ understanding of subtle actions, enhancing semantic reasoning and dense caption generation. 

This work is a first step towards building multimodal foundation models that incorporate the human skeleton modality. Future research will explore replacing SkeletonCLIP with other effective skeleton-language models.

\section*{Acknowledgements}
We thank the members of the Genius Lab at UNC Charlotte for helpful discussions. This work is supported in part by the National Science Foundation (IIS-2245652). Additionally, this material is based on research partially supported by the Chateaubriand Fellowship of the Office for Science and Technology of the Embassy of France in the United States.

\bibliography{refs}

\clearpage

\noindent \textbf{\huge Appendix}
\appendix

\section{Overview}

The Supplementary material is organized as follows:

\begin{itemize}
    \item Section \ref{sec:gradcam}: Effectiveness of SKI-VLM for ADL
    \item Section \ref{sec:ablation}: Ablations
    \begin{itemize}
        \item Section \ref{sec:txt_enc}: SkeletonCLIP Text Encoder
        \item Section \ref{sec:prtrn}: Pretraining Strategy
    \end{itemize}
    \item Section \ref{sec:lvlm_data}: LVLM Video Instruction Data Curation 
    \item Section \ref{sec:skclip}: Integrating SkeletonCLIP in VLMs
    \item Section \ref{sec:alpha}: Distillation Loss Weightage 
    
\end{itemize}

\begin{figure}[h]
  \centering
  \includegraphics[height=5.2cm, width=7.8cm]{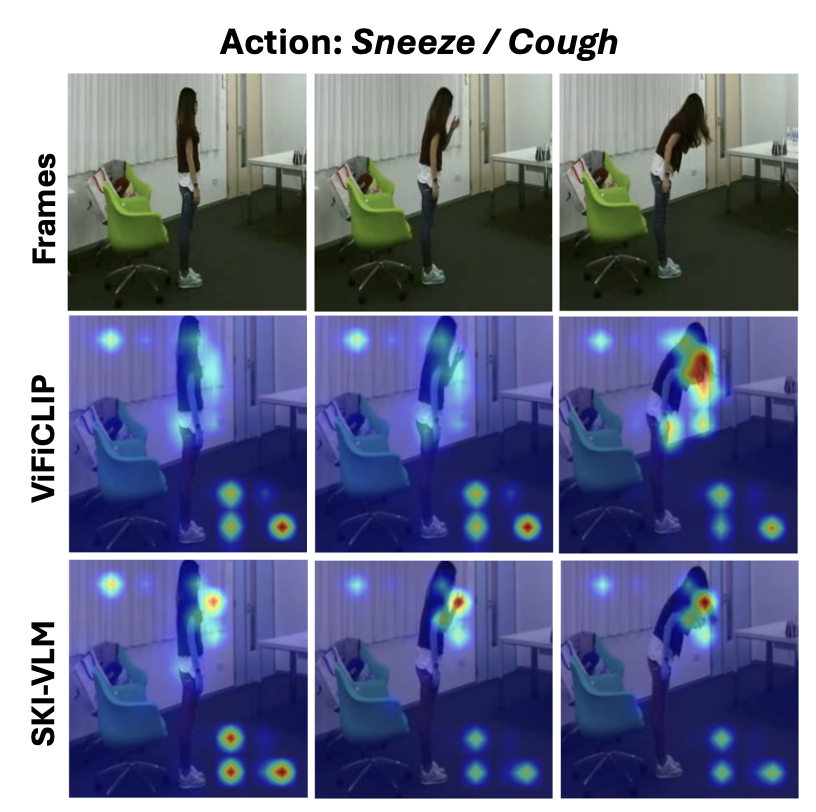}
  \caption{\textbf{Attention Map Visualization}: Comparison between ViFiCLIP and SKI-VLM. While ViFiCLIP struggles to identify the critical areas responsible for actions, SKI-VLM accurately focuses on the relevant joints, such as hands and face, for actions like \textit{Sneeze/Cough}. 
  }
  \label{fig:gradcam}
\end{figure}

\section{Effectiveness of SKI-VLM for ADL}
\label{sec:gradcam}

In Figure~\ref{fig:gradcam}, we present the attention map visualizations \footnote{The visualizations tend to emphasize static elements of the background due to absence of registers during training, as discussed in ~\cite{register_viz}. These high background values are disregarded in our analysis.} of ViFiCLIP and SKI-VLM on a sample from the  NTU dataset. These heatmaps highlight the important regions of the video frames that contributed most to the model's prediction. In the figure, the second row consists of the attention maps of the ViFiCLIP model, where the heatmaps are dispersed across large areas, indicating a broad focus. 
In contrast, the final row shows SKI-VLMs GradCAM visualization, which precisely emphasizes the relevant human joints involved in the action. 
This demonstrates that incorporating skeleton information through SKI-VLM enables a more targeted focus on the critical regions of the video compared to VifiCLIP.

\section{Ablations}
\label{sec:ablation}
\subsection{SkeletonCLIP Text Encoder}
\label{sec:txt_enc}

\begin{table}[h]
\centering
\centering
\caption{Effect of SkeletonCLIP Text Encoder on SKI-ViFiCLIP}
\scalebox{0.8}{
\begin{tabular}{lccc}
\toprule
\textbf{Method} & \textbf{NTU60} & \textbf{NTU120} & \textbf{Harmonic} \\
& \textbf{48/12} & \textbf{110/10} & \textbf{Mean} \\
\midrule
Trainable Text Encoder & \textbf{52.4} & 72.2 & 60.7\\
\rowcolor{gray!25} \textbf{Frozen Text Encoder (ours)} & 52.0 & \textbf{77.5} & \textbf{62.2}\\
\bottomrule
\end{tabular}}
\label{tab:text_enc}
\end{table}

\noindent
In this section, we evaluate the configuration of the SkeletonCLIP Text Encoder, specifically examining whether the text encoder should be trainable or frozen. As shown in Table~\ref{tab:text_enc}, SKI-ViFiCLIP achieves superior performance on average with a frozen Text Encoder compared to a trainable one. This outcome highlights the effectiveness of a frozen Text Encoder in aligning the skeleton distribution more closely with the pre-aligned text and video distributions, enhancing skeleton-text-video alignment during the SkeletonClip Distillation.

\subsection{Pretraining Strategy}
\label{sec:prtrn} 
\begin{table}[h]
\centering
\caption{Pretraining Strategy of Components of SKI-VLM}
\scalebox{0.67}{
\begin{tabular}{llccc}
\toprule
\textbf{Method} & \textbf{Pretraining} & \textbf{NTU60} & \textbf{NTU120} & \textbf{Harmonic}\\
 &  & \textbf{48/12} & \textbf{110/10} & \textbf{Mean}\\
\midrule
Online Distillation & & 50.0 & 10.0 & 16.7 \\
 & + SkeletonCLIP & 49.2 &  38.7 & 43.3 \\
 & + ViFiCLIP  & \textbf{53.0} & 69.6 & 60.2 \\
 & + SkeletonCLIP + ViFiCLIP  & 52.0 & \textbf{77.5} & \textbf{62.2}\\
\bottomrule
\end{tabular}}
\label{tab:pretraining}
\end{table}

\noindent
To validate our model’s pretraining strategy, we conducted experiments across various scenarios: no pretraining, SkeletonCLIP pretraining, ViFiCLIP pretraining, and combined SkeletonCLIP + ViFiCLIP pretraining. As shown in Table~\ref{tab:pretraining}, the model demonstrates superior performance when both SkeletonCLIP and ViFiCLIP are pretrained. This result supports our hypothesis that pretraining and aligning SkeletonCLIP are essential for it to effectively serve as a teacher model during distillation. Pretraining the skeleton encoder is critical for aligning skeleton and text representations within SkeletonCLIP. Likewise, a pretrained ViFiCLIP plays a key role in extracting discriminative video-text representations from the input videos and text prompts. 


\section{LVLM Video Instruction Data Curation}
\label{sec:lvlm_data} 

To generate the the video instruction data for training the LVLM, for each video we first crop out the person(s) performing the action. This helps in eliminating unnecessary background information in the videos. Subsequently a single frame is selected from the video and CogVLM~\cite{cogvlm} is used to generate its caption. CogVLM employs Vicuna v1.5 7B ~\cite{vicuna2023} as its primary language model and EVA2-CLIP-E~\cite{eva_CLIP} as the visual transformer encoder, with input images set at 224 × 224 pixels. 
The prompt used for captioning is - "\textit{Give a detailed description of the actions happening and describe the image, include motions and the objects interacted by the person.}"

\begin{figure*}[h]
  \centering
  \includegraphics[width=1\linewidth]{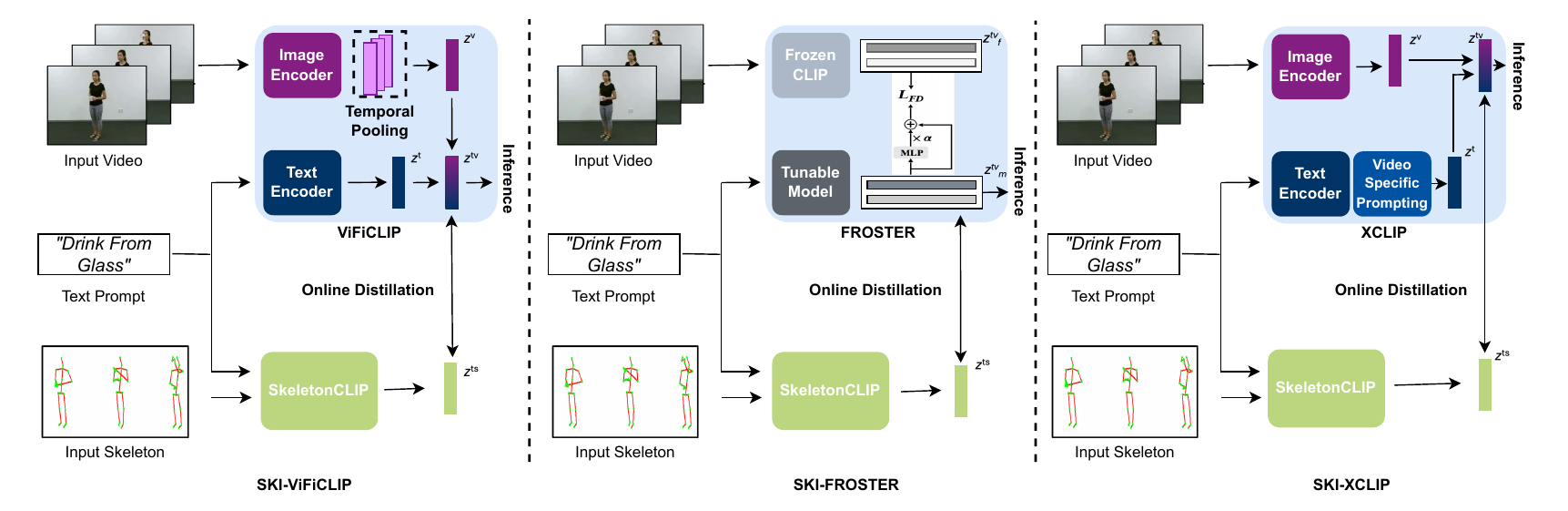}
  \caption{Illustration of how SkeletonCLIP can be easily integrated with VLMs like ViFiCLIP, FROSTER and XCLIP (\textit{left to right)}}
  \label{fig:posevidclip}
\end{figure*}

\section{Integrating SkeletonCLIP in VLMs}
\label{sec:skclip}
Figure \ref{fig:posevidclip} demonstrates SKI-VLM implemented using SkeletonCLIP as the teacher and XCLIP, ViFiCLIP, and FROSTER as the student models for performing SCD. 

\section{Distillation Loss Weightage}
\label{sec:alpha}

\begin{figure}[h]
  \centering
\includegraphics[width = 0.35\textwidth]{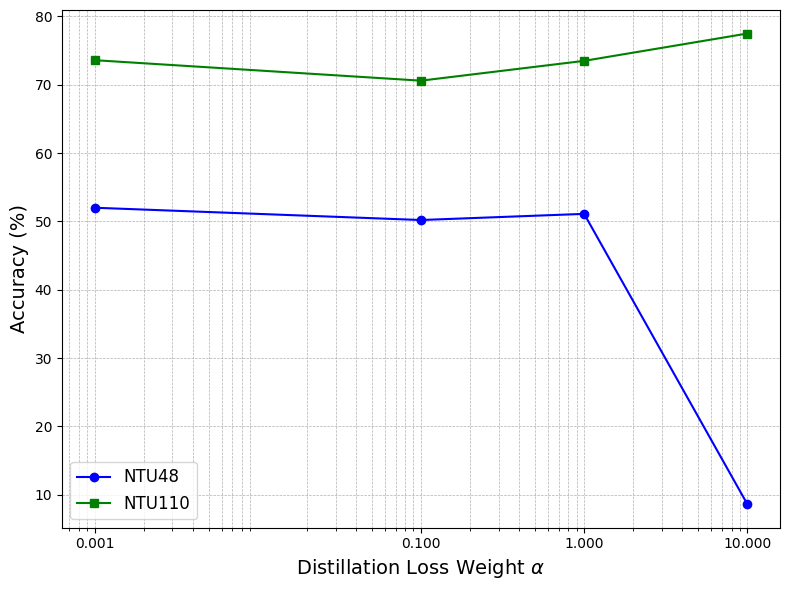}
  \caption{Impact of $\alpha$ in SKI-VLM for NTU48 and NTU110}
  \label{fig:alpha}
\end{figure}

In Figure~\ref{fig:alpha}, we report the accuracy of SKI-VLM on NTU48 and NTU110 for various values of $\alpha$. We observe that for NTU48, the optimal $\alpha$ is 0.01, while for NTU110, an $\alpha$ value of 10.0 yields the best accuracy. This indicates that in the low data regime (NTU48), placing greater emphasis on the classification loss (cross entropy) is crucial to ensure the model effectively learns from the limited data. Conversely, in the large-scale dataset (NTU110), a higher $\alpha$ value allows for more effective leveraging of the SkeletonCLIP through the distillation loss, helping the model to better generalize by aligning with the teacher model’s guidance.

\end{document}